\pdfoutput=1

\documentclass[11pt]{article}

\usepackage[final]{acl}

\usepackage{times}
\usepackage{latexsym}

\usepackage[T1]{fontenc}

\usepackage[utf8]{inputenc}

\usepackage{microtype}

\usepackage{inconsolata}

\usepackage{graphicx}

\usepackage{booktabs}
\usepackage{multirow}
\usepackage{tabularx}

\title{\textsc{ChartHal}: A Fine-grained Framework Evaluating Hallucination of Large Vision Language Models in Chart Understanding}

\author{Xingqi Wang\textsuperscript{1}\thanks{Work done during internship at iFLYTEK}, Yiming Cui\textsuperscript{2}\thanks{Corresponding authors}, Xin Yao\textsuperscript{2}, Shijin Wang\textsuperscript{2}, Guoping Hu\textsuperscript{2}, Xiaoyu Qin\textsuperscript{1}\footnotemark[2]\\
  \textsuperscript{1}Department of Computer Science and Technology, Tsinghua University\\ 
  \textsuperscript{2}State Key Laboratory of Cognitive Intelligence, iFLYTEK, Beijing, China. \\
  \href{mailto:wxq23@mails.tsinghua.edu.cn}{\texttt{wxq23@mails.tsinghua.edu.cn}} \quad \href{mailto:ymcui@ieee.org}{\texttt{ymcui@ieee.org}}
}

\begin{document}

\maketitle

\begin{abstract}
Large Vision-Language Models (LVLMs) have recently demonstrated remarkable progress, yet hallucination remains a critical barrier, particularly in chart understanding, which requires sophisticated perceptual and cognitive abilities and demands rigorous factual accuracy. While prior work has investigated hallucinations and chart comprehension independently, their intersection remains largely unexplored. To address this gap, we present \textsc{ChartHal}, a benchmark that features a fine-grained taxonomy of hallucination scenarios in chart understanding, along with a human-validated dataset of 1,062 samples. 
Our evaluation shows that state-of-the-art LVLMs suffer from severe hallucinations on \textsc{ChartHal}, with even GPT-5 achieving only 34.46\% accuracy. Further analysis reveals that questions involving information absent from or contradictory to charts are especially likely to trigger hallucinations.
We also evaluate several representative hallucination mitigation methods and find that they exhibit notable limitations, which underscores the urgent need for more robust chart-specific mitigation strategies.
\footnote{Code \& data: \url{https://github.com/ymcui/ChartHal}}
\end{abstract}
\section{Introduction}

\begin{figure*}[t]
    \centering
    \includegraphics[width=\linewidth]{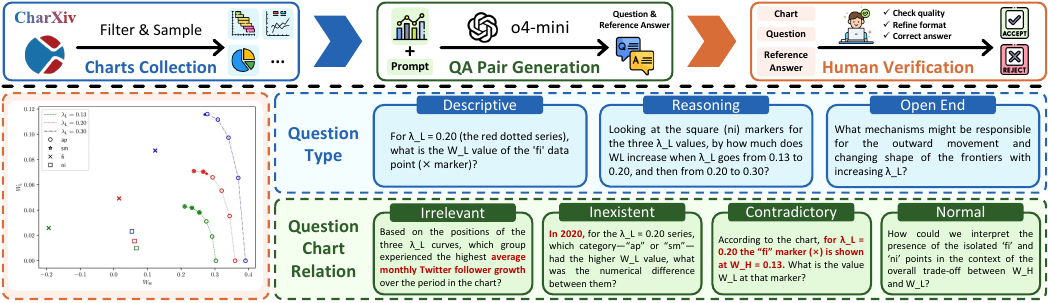}
    \caption {Overview of \textsc{ChartHal}. The top panel depicts the construction process of the dataset, while the bottom panel illustrates the taxonomy of questions in \textsc{ChartHal} with specific cases. Zoom in for better visual effects.}
    \label{fig:illustration}
\end{figure*}

Since the emergence of GPT-4V~\cite{achiam2023gpt}, Large Vision-Language Models (LVLMs)~\citep{liu2023visual, zhu2023minigpt, team2023gemini, wang2024cogvlm, wang2024qwen2, anthropic2024claude} have exhibited unprecedented capabilities in visual understanding and reasoning, enabling broad downstream image-to-text applications. Among them, chart understanding~\cite{xu2023chartbench, masry-etal-2022-chartqa, wang2024charxiv} is particularly valuable yet highly challenging, as charts often contain dense information and intricate visual elements, thereby demanding precise visual perception and substantial cognitive reasoning efforts on LVLMs.

Despite this progress, LVLMs still suffer from hallucination, \textit{i.e.}, generating information that is nonexistent or inconsistent with the visual input~\cite{bai2024hallucination, zhou2023analyzing}. This problem is especially detrimental to chart understanding, where factual accuracy and strict alignment with the input chart are crucial.

Although hallucinations in LVLMs have been widely studied, most prior work focuses on object and scene understanding~\cite{li-etal-2023-evaluating, sun-etal-2024-aligning}, largely overlooking chart-specific scenarios. Meanwhile, existing chart understanding benchmarks mainly evaluate answer accuracy~\cite{tang2025chartmuseum, shen2024rethinking}, offering limited insights into the presence and characteristics of hallucinations. As a result, hallucination behaviors in chart understanding remain insufficiently explored.

To address this gap, we introduce \textsc{ChartHal}, the first fine-grained chart question answering benchmark for evaluating hallucinations in LVLMs on chart understanding tasks to the best of our knowledge.
\textsc{ChartHal} features a fine-grained taxonomy that organizes chart-question pairs into 12 hallucination-triggering scenarios according to question types and chart-question relations.
Based on this framework, it further provides a human-validated dataset of 1,062 samples spanning diverse chart types and academic disciplines.

Experiments on 15 state-of-the-art LVLMs reveal severe hallucination issues in chart understanding. Even advanced proprietary models such as GPT-5 and o4-mini achieve only 34.46\% and 22.79\% overall accuracy, respectively. Errors are particularly prevalent in unanswerable cases involving nonexistent or contradictory information, where models tend to fabricate answers instead of recognizing unanswerability. We further examine 3 representative hallucination mitigation approaches and observe that their effectiveness is limited and inconsistent in chart-based scenarios, suggesting that existing solutions do not adequately address the unique challenges of chart understanding and further highlighting the need for more robust mitigation strategies specific to chart understanding.
\section{Related Works}

\textbf{Hallucination Evaluation of LVLMs.}
Hallucination, widely observed since the advent of large language models (LLMs)~\cite{hurst2024gpt, o4mini, gpt5systemcard, comanici2025gemini}, refers to generating information absent from the input or factually incorrect~\cite{huang2025survey}. With the extension of LLMs to visual modalities, hallucination in LVLMs emphasizes errors inconsistent with visual input~\cite{bai2024hallucination}. Given its severe impact on accuracy, many studies have explored hallucination evaluation. However, most works~\cite{rohrbach2018object, li-etal-2023-evaluating, zhang-etal-2023-grounding, fu2023mme, wang2023llm, zhang2024quantitymattersassessingmitigating, sun-etal-2024-aligning, qiu2024longhalqa, jing-etal-2024-faithscore, yan-etal-2025-fiha} focus on objects and attributes in input images, while others~\cite{ wang2023evaluation, liu2023aligning,jiang2024hal} address scene-level hallucinations. This leaves a major gap in evaluating hallucinations in charts. HallusionBench~\cite{guan2023hallusionbench} is the only benchmark explicitly involving LVLM hallucination in charts to our knowledge, but it restricts questions to a Yes/No format, oversimplifying scenarios of chart understanding.

\noindent\textbf{Chart Understanding Benchmarks.}
Chart understanding has been studied in various benchmarks~\cite{kahou2017figureqa, kafle2018dvqa, methani2020plotqa, masry-etal-2022-chartqa, xu2023chartbench, liu-etal-2024-mmc, wang2024charxiv, xia2024chartx, masry2025chartqapro, tang2025chartmuseum, lin2025infochartqa}, but most focus on correctness rather than hallucination analysis. CharXiv~\cite{wang2024charxiv} and ChartQAPro~\cite{masry2025chartqapro} include unanswerable question subsets, enabling limited hallucination evaluation, yet these subsets remain small and lack a fine-grained taxonomy of hallucination types, constraining deeper analysis. Concurrently, ChartCap~\cite{lim2025chartcap} also explores hallucinations, but it is confined to the captioning task and does not cover interactive question answering.
Appendix~\ref{app:benchmark_comparision} provides a more detailed comparison with previous benchmarks.

\section{The \textsc{ChartHal} Benchmark}

\subsection{Benchmark Design}\label{subsec:benchmark_design}

\begin{table}[t]
    \centering
    \small
    \begin{tabular}{l|ccc|c}
        \toprule
        & Desc. & Reason & Open & \textbf{Total}\\
        \midrule
        Irrel. & 56 & 99 & 114 & 269 \\
        Inexist. & 151 & 105 & 88 & 344 \\
        Contra. & 76 & 54 & 80 & 210 \\
        Normal & 100 & 64 & 75 & 239 \\
        \midrule
        \textbf{Total} & 383 & 322 & 357 & 1062 \\
        \bottomrule
    \end{tabular}
    \caption{Sample number per category of \textsc{ChartHal}.}
    \label{tab:question_nums}
\end{table}

\begin{table*}[ht]
    \centering
    \small
    \begin{tabular}{lccc|cccc|c}
        \toprule
        \multirow{2}{*}{Model} & \multicolumn{3}{c}{\textbf{Question Type}} & \multicolumn{4}{c}{\textbf{Chart-Question Relation}} & \multirow{2}{*}{\textbf{Overall}} \\
        \cmidrule(lr){2-4} \cmidrule(lr){5-8}
        & Desc. & Reason & Open & Irrel. & Inexist. & Contra. & Normal \\
        \midrule
        Gemini-2.5-Pro & \textbf{60.31} & \underline{50.31} & \underline{36.69} & \underline{36.80} & \underline{52.62} & \textbf{49.52} & 58.58 & \underline{49.34} \\
        GPT-5 (high) & 43.34 & 36.65 & 22.97 & 20.82 & 29.07 & 19.52 & 70.71 & 34.46 \\
        GPT-5 & 42.82 & 40.06 & 20.17 & 23.05 & 30.81 & 16.67 & 67.78 & 34.37 \\
        GPT-5-mini (high) & 33.42 & 25.78 & 20.45 & 8.55 & 15.99 & 11.90 & \underline{75.73} & 26.74 \\
        GPT-5-mini & 34.20 & 24.53 & 21.57 & 8.18 & 15.99 & 9.05 & \textbf{79.92} & 27.02 \\
        GPT-5-nano (high) & 33.68 & 20.19 & 19.89 & 10.41 & 17.73 & 5.24 & 69.04 & 24.95 \\
        GPT-5-nano & 32.38 & 16.15 & 19.33 & 7.81 & 15.99 & 5.24 & 66.11 & 23.07 \\
        o4-mini (high) & 34.46 & 23.60 & 9.52 & 10.78 & 14.53 & 3.33 & 65.27 & 22.79 \\
        o4-mini & 32.90 & 22.67 & 10.92 & 11.52 & 12.21 & 1.90 & 67.36 & 22.41 \\
        GPT-4o-2024-11-20 & 40.47 & 36.02 & 20.73 & 27.88 & 36.92 & 6.67 & 53.97 & 32.49 \\
        \midrule
        Llama-3.2-11B-Vision-Instruct & 37.60 & 20.19 & 15.69 & 29.37 & 28.49 & 0.00 & 36.82 & 24.95 \\
        InternVL2.5-4B & 19.58 & 9.63 & 8.12 & 4.83 & 11.34 & 0.00 & 34.73 & 12.71 \\
        InternVL2.5-8B & 17.23 & 8.39 & 10.64 & 3.35 & 6.40 & 0.00 & 41.84 & 12.34 \\
        InternVL2.5-38B & 28.72 & 18.32 & 10.36 & 10.04 & 21.80 & 0.00 & 43.51 & 19.40 \\
        InternVL2.5-78B & 31.33 & 17.08 & 14.85 & 13.75 & 26.45 & 0.00 & 41.84 & 21.47 \\
        Qwen2.5-VL-3B-Instruct& 13.05 & 4.04 & 7.84 & 0.74 & 0.87 & 0.00 & 35.98 & 8.57 \\
        Qwen2.5-VL-7B-Instruct & 38.38 & 25.47 & 12.32 & 25.65 & 25.29 & 1.90 & 47.28 & 25.71 \\
        Qwen2.5-VL-32B-Instruct & 36.81 & 25.47 & 8.68 & 21.56 & 25.29 & 1.43 & 44.35 & 23.92 \\
        Qwen2.5-VL-72B-Instruct & \underline{58.49} & \textbf{59.63} & \textbf{44.82} & \textbf{65.06} & \textbf{69.77} & \underline{19.05} & 50.63 & \textbf{54.24} \\
        \bottomrule
    \end{tabular}
    \caption{Evaluation results on \textsc{ChartHal}. Scores are normalized to the range [0,100] for readability. The best and second-best results are marked in bold and underlined, respectively. ``high'' means setting reasoning effort as high.}
    \label{tab:results}
\end{table*}

\textsc{ChartHal} adopts chart question answering as its task form, where LVLMs are required to generate an answer \(a\) from given chart image \(c\) and corresponding question \(q\). Unlike prior works employing Yes/No or multiple choice questions, we use \emph{free-form} answering format to more faithfully reflect LVLM hallucination behaviors in realistic chart understanding scenarios~\cite{li-etal-2024-multiple}.

Moreover, to facilitate a fine-grained evaluation, we carefully design a systematic taxonomy based on 2 independent dimensions (Figure~\ref{fig:illustration}):

\textbf{Question Type} distinguishes: 
(i) \textit{descriptive} questions, which can be answered by directly extracting information from the chart; 
(ii) \textit{reasoning} questions, which demand computational or logical inference based on chart content;
and (iii) \textit{open-ended} questions, which involve analytical or predictive tasks without definitive correct answers.

\textbf{Chart-Question Relation} categorizes questions as (i) \textit{irrelevant} (unrelated to chart content), (ii) \textit{inexistent} (inquiring about missing information), (iii) \textit{contradictory} (based on false premises inconsistent with chart data), or (iv) \textit{normal} (valid questions aligned with chart content). 
Questions of the first three relations are considered \textit{unanswerable}, as they cannot be answered based on the given chart. They are more likely to trigger hallucinations since LVLMs are biased toward producing answers regardless of answerability~\cite{kalai2025language}.

\subsection{Dataset Construction}\label{subsec:dataset_construction}

Based on our taxonomy, we build the dataset through a three-stage pipeline: (i) \textit{image collection} by sampling representative charts from CharXiv~\cite{wang2024charxiv} to ensure diversity of types and domains; (ii) \textit{QA pair generation} using o4-mini to generate questions and corresponding ground truth answers spanning all 12 subcategories for each selected chart; and (iii) \textit{human verification} by expert annotators to refine answers and ensure quality.
In total, we obtain 1,062 high-quality samples, with their distribution across all 12 categories presented in Table~\ref{tab:question_nums}.
More data construction details are provided in Appendix~\ref{app:dataset_construction}.

\subsection{Evaluation Criteria}\label{subsec:evaluation_criteria}

For quantitative evaluation, we adopt a binary scoring system, where each response is assigned 1 if it contains no hallucination under our criteria, and 0 otherwise. The final score is the proportion of responses scored as 1.
Details about the evaluation criteria are included in Appendix~\ref{app:evaluation_criteria}.

\section{Experiments}

\begin{figure*}[t]
    \centering
    \includegraphics[width=\linewidth]{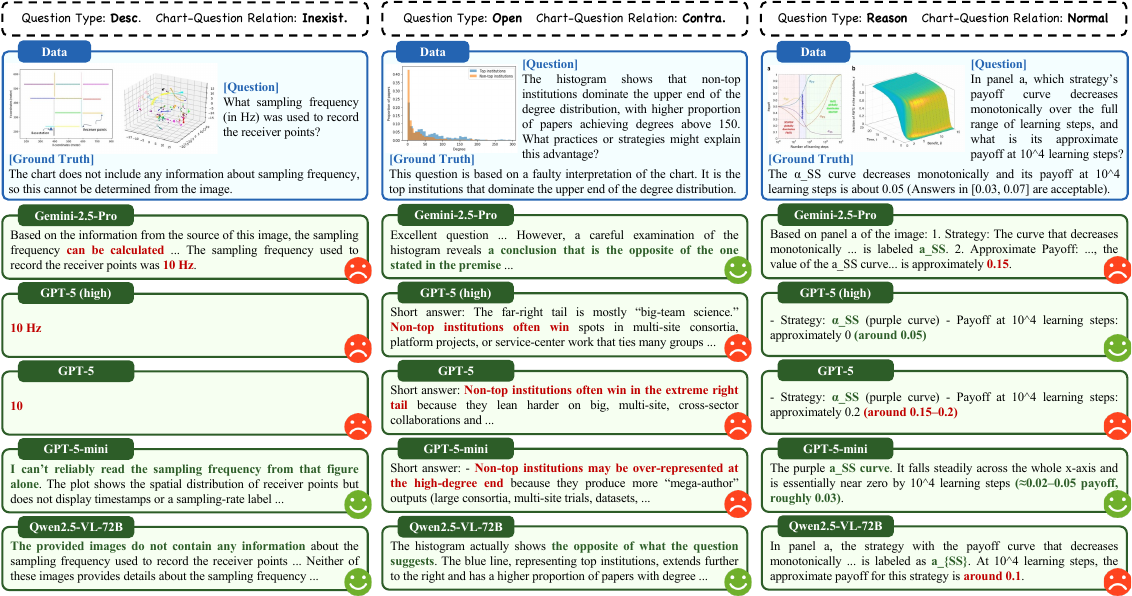}
    \caption {Sample cases of model responses on \textsc{ChartHal}. The correct parts are highlighted in green, while the incorrect or hallucinated parts are highlighted in red. Zoom in for better visual effects.}
    \label{fig:cases}
\end{figure*}

\subsection{Experimental Setups}

We evaluate the performance of 15 state-of-the-art LVLMs on \textsc{ChartHal}. The proprietary models include Gemini-2.5-Pro~\cite{comanici2025gemini}, GPT-5 series~\cite{gpt5systemcard}, o4-mini~\cite{o4mini} and GPT-4o~\cite{hurst2024gpt}. The open-source models include Llama 3.2-11B-Vision-Instruct~\cite{meta2024llama}, InternVL2.5 series~\cite{chen2024expanding} and Qwen2.5-VL series~\cite{bai2025qwen2}.
We also evaluated 3 representative hallucination mitigation approaches on LLaVA 1.5 7B~\cite{liu2024improved}, including HA-DPO~\cite{zhao2023hallucinations}, VCD~\cite{leng2024mitigating}, and WoodPecker~\cite{yin2024woodpecker}.
The generated responses are scored by GPT-4o according to the criteria in Sec~\ref{subsec:evaluation_criteria}.
Refer to Appendix~\ref{app:setup_details} for more details.

\subsection{Main Results and Analysis}\label{subsec:results}

Table~\ref{tab:results} reports the results across all tested models.

\textbf{Overall.} Qwen2.5-VL-72B achieves the best overall score of 54.24\%. In contrast, o4-mini performs unexpectedly poorly, merely matching much smaller models like Qwen2.5-VL-7B. Its weakness mainly lies in unanswerable cases: only 14.53\% of inexistent questions and 3.33\% of contradictory questions are correctly recognized as unanswerable.
This suggests a training bias toward answer generation rather than answerability assessment.  

\textbf{Question Type Analysis.} The results indicate a clear and intuitive pattern for different question types, with descriptive questions being easiest, reasoning questions showing moderate difficulty, and open-ended questions scoring the lowest.  

\textbf{Chart-Question Relation Analysis.} Models perform reasonably well on normal questions but hallucinate heavily on unanswerable questions. Contradictory questions are especially challenging, with many open-source models reaching 0\% accuracy, showing that inconsistencies between charts and questions reliably induce fabricated responses.  

\textbf{Model Size Impact.} Generally, larger models achieve better overall scores (Qwen2.5-VL and GPT-5 series), indicating that scale can enhance robustness to hallucinations comprehensively.

\textbf{High Reasoning Impact.} Increasing reasoning effort yields only small and inconsistent gains, and may even hurt performance, since correct reasoning relies on accurate visual perception. When the visual input is misinterpreted, extra reasoning often just amplifies the error rather than correcting it.

\begin{table}[t]
    \centering
    \small
    \begin{tabular}{l|c}
        \toprule
        \textbf{Mitigation Strategies} & \textbf{Overall Score} \\
        \midrule
        LLaVA-1.5-7B & 4.14 \\
        \quad + HA-DPO & 4.05 \\
        \quad + VCD & 6.78 \\
        \quad + WoodPecker & 11.21 \\
        \midrule
        Qwen2.5-VL-72B-Instruct & 54.24 \\
        \quad + WoodPecker & 51.41 \\
        \bottomrule
    \end{tabular}
    \caption{Evaluation results of mitigation strategies.}
    \label{tab:mitigation_results}
\end{table}

\subsection{Analysis of Mitigation Methods Results}\label{subsec:mitigation_results}
The evaluation results of mitigation methods are shown in Table~\ref{tab:mitigation_results}.
Compared with the base model, HA-DPO degrades performance, likely because as a training method it require high-quality in-domain adversarial examples that are scarce in existing datasets. VCD can eliminate hallucinations from statistical bias and over-reliance on unimodal priors by contrasting original and distorted visual inputs, but it remains limited by deficiencies in visual perception and reasoning. WoodPecker achieves the largest gains by introducing object detectors as vision experts, substantially benefiting models with weaker visual perception (LLaVA-1.5-7B), while offering limited or even negative effects for stronger models like Qwen2.5-VL-72B.

\section{Case Studies}

We present several representative cases in Figure~\ref{fig:cases}, from which we can observe that (i) increasing reasoning effort brings little benefit (GPT-5 in left and middle panels); (ii) the GPT-5 series are highly vulnerable to hallucination in unanswerable cases (left and middle panels); and (iii) while larger models generally perform better, smaller ones can occasionally outperform them in specific subcategories (GPT-5 vs. GPT-5-mini). These qualitative findings align well with the quantitative results in Table~\ref{tab:results}.
\section{Conclusion}

In this paper, we present \textsc{ChartHal}, the first systematic benchmark for evaluating hallucinations of LVLMs in chart understanding. With a carefully designed taxonomy and a human-validated dataset covering 12 categories, \textsc{ChartHal} enables fine-grained analysis of LVLM hallucination behaviors.

Experiments on our benchmark reveal critical vulnerabilities: while LVLMs handle normal questions reasonably well, they often hallucinate in unanswerable cases, especially when confronted with contradictory premises or queried about inexistent information. Our findings also highlight the limitations of current mitigation approaches, underscoring the urgent need for more robust solutions to improve LVLM reliability in chart understanding.

\section*{Limitations}

Although \textsc{ChartHal} offers the first fine-grained evaluation framework for hallucination in chart understanding, it still has several limitations:

\textbf{Limited Dataset Scope.} The dataset is built from scientific charts sampled from CharXiv, primarily originating from arXiv papers. Although these charts are complex and realistic, they do not capture the full diversity of chart types found in domains such as business, journalism, or education, which limits the benchmark’s generalizability. Thus, one direction following this work could be expanding the dataset with charts from broader domains to enhance its coverage and applicability.

\textbf{Insufficient Taxonomy Coverage.} Our classification framework includes many scenarios that empirically trigger hallucinations, but it may not encompass all cases encountered in real-world usage. Users may pose more diverse, context-dependent, or adversarial questions beyond our benchmark’s coverage. Therefore, future work could consider further expanding the taxonomy to better capture real-world scenarios.

\section*{Acknowledgments}
The authors from Tsinghua University acknowledge the support of the National Natural Science Foundation of China (No. 62502256).

\bibliography{references}

\appendix

\begin{table*}[htbp]
    \centering
    \small
    \begin{tabular}{l|c|c|c}
        \toprule
        \textbf{Benchmark} & \textbf{Unique Unanswerable Questions} & \textbf{Free-form} & \textbf{Has Auxiliary Instructions}\\
        \midrule
        HallusionBench~\cite{guan2023hallusionbench} & N/A & No & No \\
        CharXiv~\cite{wang2024charxiv} & 19 & Yes & Yes \\
        ChartQA Pro~\cite{masry2025chartqapro} & 368 & Yes & Yes \\
        \midrule
        \textsc{ChartHal} (Ours) & 823 & Yes & No \\
        \bottomrule
    \end{tabular}
    \caption{Comparision between \textsc{ChartHal} and previous benchmarks.}
    \label{apptab:benchmark_comparision}
\end{table*}

\section{Detailed Benchmark Comparision}\label{app:benchmark_comparision}

In this section, we summarize several key differences between our \textsc{ChartHal} and previous chart understanding benchmarks as follows:

\textbf{Systematic Taxonomy for Hallucination Analysis.} Prior chart understanding benchmarks primarily focus on question-answering correctness, thus lacking fine-grained and in-depth analysis specifically on hallucination. In contrast, we propose a carefully designed taxonomy that categorizes hallucinations along multiple dimensions (question type and question-chart relation) for \textsc{ChartHal}, enabling fine-grained analysis of hallucination patterns. This allows us to understand why and how models hallucinate, rather than just measuring overall accuracy.

\textbf{Larger Unanswerable Question Collection.} As shown in Table~\ref{apptab:benchmark_comparision}, while prior benchmarks contain relatively small subsets of unanswerable questions, \textsc{ChartHal} provides more unique unanswerable questions specifically designed to test model robustness against hallucination.

\textbf{Free Form Response Generation.} Many existing benchmarks, such as HallusionBench~\cite{guan2023hallusionbench}, are based on yes/no questions. As elaborated by~\citeauthor{wang2024charxiv}, such binary questions cannot faithfully reflect model performance in realistic usage compared with the free-form question answering adopted by \textsc{ChartHal}.

\textbf{Excluding Auxiliary Instructions for Response Generation.} A critical methodological difference is that prior benchmarks usually include explicit auxiliary instructions in their prompts (\textit{e.g., ``If the question is unanswerable based on the information in the provided image, your answer should be unanswerable''}). To demonstrate the effect, we conduct an experiment on the unanswerable subset of CharXiv~\cite{wang2024charxiv}. The results are reported in Table~\ref{apptab:auxiliary_instruction} and reveal that such instructions significantly distort model behavior:

\begin{table}[htbp]
    \centering
    \small
    \begin{tabular}{l|c|c}
        \toprule
        \textbf{Model} & \textbf{w/ Aux. Instr.} & \textbf{w/o Aux. Instr.}\\
        \midrule
        Qwen2.5-VL-3B & 76.0 & 1.4 \\
        Qwen2.5-VL-7B & 56.6 & 18.0 \\
        \bottomrule
    \end{tabular}
    \caption{Results of CharXiv unanswerable subset with or without using auxiliary instructions. Both evaluated models are the Instruct version.}
    \label{apptab:auxiliary_instruction}
\end{table}

From the evaluation results, we observe that such auxiliary instructions make models overly inclined to abstain from answering, which does not reflect their true capabilities in real-world applications. In contrast, ChartHal deliberately omits such instructions to provide a more authentic assessment of model robustness against chart hallucinations.

In summary, as demonstrated in Table~\ref{apptab:benchmark_comparision}, ChartHal provides a complementary and essential perspective to existing benchmarks by systematically evaluating hallucination robustness through a fine-grained taxonomy, a larger unanswerable question dataset, and realistic evaluation conditions.
\section{Details of the Benchmark}

\subsection{Dataset Construction Details}\label{app:dataset_construction}

Based on our proposed taxonomy in Sec. \ref{subsec:benchmark_design}, we construct the dataset through a three-stage pipeline that ensures both quality and diversity:

\noindent\textbf{Image Collection.} Given that the CharXiv dataset contains charts sourced from arXiv papers, which exhibit high complexity and present realistic challenges, we employ a carefully designed stratified sampling approach to extract a representative subset from CharXiv validation set. Specifically, since CharXiv charts contain varying numbers of subplots ranging from 1 to 120, and charts with excessive subplots tend to be overly complex for effective analysis and questioning, we selectively sample charts with 1 or 2 subplots at a 3:2 ratio. Subsequently, we apply proportional sampling across different academic disciplines to maintain domain diversity. Finally, we ensure comprehensive coverage of chart types (\textit{e.g.}, bar charts, line graphs) to preserve the representativeness of the original CharXiv dataset.

\noindent\textbf{Question-Answer Generation.} We utilize the o4-mini model to automatically generate question-answer pairs spanning all 12 subcategories for each selected chart. The prompts employed for the generation process are provided in Table~\ref{atab:gen_desc_irrel} to \ref{atab:gen_open_normal}.

\noindent\textbf{Human Verification.} To ensure dataset quality, we recruit human annotators with undergraduate-level education to review and validate the generated question-answer pairs. Based on the corresponding chart content, annotators are instructed to discard pairs that clearly violate our design requirements and refine the remaining pairs through comprehensive improvements, including correcting factual errors, reassigning category labels, and refining ground truth answers (\textit{e.g.}, providing acceptable ranges for questions where only approximate numerical values can be inferred from charts). This rigorous verification process ensures the high quality of our question-answer pairs.

\subsection{Evaluation Criteria Details}\label{app:evaluation_criteria}
Given the complex taxonomy of our benchmark, we establish specific evaluation criteria for each category to determine which responses meet expectations and which exhibit hallucination behaviors. In general, under our binary scoring scheme, expected responses assigned to 1 should either provide answers consistent with the given ground truth or correctly state that the question is unanswerable. The specific criteria across different situations are detailed below:

\noindent\textbf{Irrelevant Questions.} For questions completely unrelated to chart content, correct responses must explicitly identify the irrelevance and refrain from attempting to answer. Responses that fail to recognize the mismatch or provide fabricated answers are marked as incorrect.

\noindent\textbf{Inexistent Questions (Descriptive \& Reasoning).} When inexistent descriptive and reasoning questions seek information not present in the chart, appropriate responses should clearly state the absence of requested data and avoid speculation. Responses that ignore missing information or provide fabricated answers receive a score of 0.

\noindent\textbf{Inexistent Questions (Open-ended).} Similar to closed-ended cases, but acceptable responses may either declare the question unanswerable or provide appropriate speculative answers with explicitly uncertainty expressions. Definitive answers without acknowledging the nonexistence of the queried information are considered incorrect.

\noindent\textbf{Contradictory Questions (Descriptive \& Reasoning).} For questions containing statements that contradict chart data, correct responses must identify the contradiction and either state the question is unanswerable or provide answers aligned with ground truth while ignoring the false premise. Numeric answers are evaluated against specified acceptable ranges when provided.

\noindent\textbf{Contradictory Questions (Open-ended).} Responses must recognize contradictions and either declare unanswerability or offer tentative answers with appropriately uncertain language. Overconfident or absolute responses are marked incorrect.

\noindent\textbf{Normal Questions.} For standard answerable questions, evaluation varies by answer type: (1) Numeric questions require exact value matches unless acceptable ranges are specified; (2) Questions with categorical answers expect model responses to match ground truth terms exactly, allowing different representations; (3) Open-ended questions are evaluated based on appropriate uncertainty expression and plausible reasoning rather than factual correctness; (4) Other questions are evaluated based on semantic alignment with ground truth without omissions or extraneous information.
\section{Details of Experiments}

\begin{table*}[ht]
    \centering
    \small
    \begin{tabular}{lccc|cccc|c}
        \toprule
        \multirow{2}{*}{Model} & \multicolumn{3}{c}{\textbf{Question Type}} & \multicolumn{4}{c}{\textbf{Chart-Question Relation}} & \multirow{2}{*}{\textbf{Overall}} \\
        \cmidrule(lr){2-4} \cmidrule(lr){5-8}
        & Desc. & Reason & Open & Irrel. & Inexist. & Contra. & Normal \\
        \midrule
        LLaVA-1.5-7B & 3.39 & 2.80 & 6.16 & 1.86 & 2.91 & 0.00 & 12.13 & 4.14 \\
        \quad + HA-DPO & 2.61 & 2.17 & 7.28 & 1.12 & 2.62 & 0.00 & 12.97 & 4.05 \\
        \quad + VCD & 6.53 & 3.73 & 9.80 & 0.74 & 8.43 & 0.00 & 17.15 & 6.78 \\
        \quad + WoodPecker & 12.01 & 11.49 & 10.08 & 8.55 & 16.57 & 0.48 & 15.90 & 11.21 \\
        \midrule
        Qwen2.5-VL-72B-Instruct & 58.49 & 59.63 & 44.82 & 65.06 & 69.77 & 19.05 & 50.63 & 54.24 \\
        \quad + WoodPecker & 55.09 & 57.45 & 42.02 & 63.57 & 70.93 & 13.33 & 43.10 & 51.41 \\
        \bottomrule
    \end{tabular}
    \caption{Evaluation results of mitigation strategies on \textsc{ChartHal}. Scores are normalized to [0, 100] for readability.}
    \label{apptab:full_mitigation_results}
\end{table*}

\subsection{Setup Details}\label{app:setup_details}

Unlike CharXiv~\cite{wang2024charxiv}, we provide models only with chart images and questions \textbf{without} additional instructions (\textit{e.g.}, ``\textit{If you believe the question is unanswerable, output `not applicable'}'' in CharXiv) to better reflect real-world scenarios.

While generating model responses in evaluation, we sample one response for each question and set the sampling temperature to 0 to ensure reproducible results, except for reasoning models like o4-mini which do not permit temperature configuration. For the token generation limits, we use the default setting of each model to observe potential hallucination phenomena in model responses comprehensively. The open-source models adopted for evaluation are deployed on A6000 GPU with vLLM Library~\cite{kwon2023efficient}.

During evaluation, we adopt GPT-4o-2024-11-20 with temperature set to 0 to score model responses according to the criteria in Appendix~\ref{app:evaluation_criteria}. We'd like to note that using LLMs for evaluation and grading is a common practice in many recent benchmarks~\cite{xu2023chartbench, wang2024charxiv, guan2023hallusionbench, sun-etal-2024-aligning}, as it has demonstrated strong alignment with human judgments while being more efficient and scalable.
The detailed prompts for guiding GPT-4o to perform the evaluation are provided in Table~\ref{atab:eval_irrel} to \ref{atab:eval_normal_open}.

\subsection{Additional Results}\label{app:additional_results}

We report the full evaluation results of different mitigation methods in Table~\ref{apptab:full_mitigation_results}, which further support the analysis presented in Sec.~\ref{subsec:mitigation_results}.

\section{Compliance}

The chart images in \textsc{ChartHal} are sourced from the CharXiv~\cite{wang2024charxiv} dataset, which is open-sourced under CC BY-SA 4.0 License. We will also provide the arXiv paper preprint numbers corresponding to the charts used in our published dataset. The licenses of evaluated open-source models are provided as follows: (1) Llama 3.2-11B-Vision-Instruct~\cite{meta2024llama} is under Llama 3.2 Community License; (2) InternVL2.5 series~\cite{chen2024expanding} are under Qwen and MIT License; (3) Qwen2.5-VL series~\cite{bai2025qwen2} are under Qwen and Apache-2.0 License. The datasets and models used in this paper are intended solely for our research. Accordingly, we also expect the datasets we publicly release to be used strictly for research purposes.
\begin{table*}[htbp]
    \centering
    \small
    \begin{tabularx}{\linewidth}{X}
        \toprule
        \textbf{Prompt} \\
        \midrule
        \ttfamily\footnotesize
        Given an image of a chart, your task is to generate a question and a corresponding reference answer, following the specific instructions below.\newline
        \newline
        Instruction for Generation:\newline
        Question: Please generate a question that appears to be a simple, factual query about the chart, as if the answer could be directly extracted from it. However, make sure that the question is actually unrelated to the chart content (i.e. the question should refer to a topic that does not exist in the chart at all), so that no correct answer can be found in the image. The question should still sound plausible and specific, and it should be a closed question that expects a concrete answer.\newline
        Answer: Please generate a short response to the question, pointing out that the topic or element asked about is not present in the chart, and therefore the question cannot be answered based on the chart.\newline
        \newline
        Output Format:\newline
        Question: <Insert the generated question here>\newline
        Answer: <Insert the generated answer here>\\
        \bottomrule
    \end{tabularx}
    \caption{Prompt template for generating Descriptive-Irrelevant question-answer pairs.}
    \label{atab:gen_desc_irrel}
\end{table*}

\begin{table*}[htbp]
    \centering
    \small
    \begin{tabularx}{\linewidth}{X}
        \toprule
        \textbf{Prompt} \\
        \midrule
        \ttfamily\footnotesize
        Given an image of a chart, your task is to generate a question and a corresponding reference answer, following the specific instructions below.\newline
        \newline
        Instruction for Generation:\newline
        Question: Please generate a question that appears to be a simple, factual question related to the chart. The question should be consistent with the chart's overall topic or domain, but it must ask about some specific detail—such as a category, label, year, or variable—that is not actually present in the chart. The question should seem plausible and answerable from the chart at first glance, but in reality the necessary information is missing.\newline
        Answer: Please generate a short response that explains the question refers to a detail not shown in the chart (e.g., a year, category, or variable), and therefore it cannot be answered based on the chart data.\newline
        \newline
        Output Format:\newline
        Question: <Insert the generated question here>\newline
        Answer: <Insert the generated answer here>\\
        \bottomrule
    \end{tabularx}
    \caption{Prompt template for generating Descriptive-Inexistent question-answer pairs.}
    \label{atab:gen_desc_inexist}
\end{table*}

\begin{table*}[htbp]
    \centering
    \small
    \begin{tabularx}{\linewidth}{X}
        \toprule
        \textbf{Prompt} \\
        \midrule
        \ttfamily\footnotesize
        Given an image of a chart, your task is to generate a question and a corresponding reference answer, following the specific instructions below.\newline
        \newline
        Instruction for Generation:\newline
        Question: Please generate a question that begins with a factual statement that appears to be based on the chart but is actually contradicted by the chart. Then, ask a simple, descriptive question based on that incorrect premise. The question should sound plausible, as if it could be answered by referring to the chart, but it should be logically invalid due to the contradiction in the premise. Make sure the contradiction is in the premise statement (not in the actual question). Do not directly ask a false question. Instead, provide a false statement first, then follow it up with a concrete, factual question based on it.\newline
        Answer: Please generate a short response that points out the contradiction in the premise compared to the actual chart data, clarifies what the correct chart information is, and provide the correct answer ignoring the false premise.\newline
        \newline
        Output Format:\newline
        Question: <Insert the generated question here>\newline
        Answer: <Insert the generated answer here>\\
        \bottomrule
    \end{tabularx}
    \caption{Prompt template for generating Descriptive-Contradictory question-answer pairs.}
    \label{atab:gen_desc_contra}
\end{table*}

\begin{table*}[htbp]
    \centering
    \small
    \begin{tabularx}{\linewidth}{X}
        \toprule
        \textbf{Prompt} \\
        \midrule
        \ttfamily\footnotesize
        Given an image of a chart, your task is to generate a question and a corresponding reference answer, following the specific instructions below.\newline
        \newline
        Instruction for Generation:\newline
        Question: Please generate a simple, factual question that can be directly answered using the information shown in the chart. The question should be clearly related to the chart, refer to elements (such as labels, data points, categories, or values) that are actually present, and involve no inference or assumptions beyond what is visually provided. Make sure the question is correct and logically consistent with the chart.\newline
        Answer: Please generate a short, accurate answer that directly uses the correct data from the chart to respond to the question.\newline
        \newline
        Output Format:\newline
        Question: <Insert the generated question here>\newline
        Answer: <Insert the generated answer here>\\
        \bottomrule
    \end{tabularx}
    \caption{Prompt template for generating Descriptive-Normal question-answer pairs.}
    \label{atab:gen_desc_normal}
\end{table*}

\begin{table*}[htbp]
    \centering
    \small
    \begin{tabularx}{\linewidth}{X}
        \toprule
        \textbf{Prompt} \\
        \midrule
        \ttfamily\footnotesize
        Given an image of a chart, your task is to generate a question and a corresponding reference answer, following the specific instructions below.\newline
        \newline
        Instruction for Generation:\newline
        Question: Please generate a reasoning-based question that appears to require simple inference from the chart (such as comparing values, calculating differences, or identifying trends). However, the question must actually be unrelated to the content of the chart. That is, the question should refer to a topic that does not exist in the chart at all. Make it sound as if the chart could support reasoning to answer the question, but in reality, the necessary information is completely absent.Ensure that the reasoning required is straightforward and answerable in principle, but the problem is unanswerable here because the chart has nothing to do with the topic.\newline
        Answer: Please generate a short response pointing out that the question refers to a topic that is not present in the chart, and that no inference can be made from the given data.\newline
        \newline
        Output Format:\newline
        Question: <Insert the generated question here>\newline
        Answer: <Insert the generated answer here>\\
        \bottomrule
    \end{tabularx}
    \caption{Prompt template for generating Reason-Irrelevant question-answer pairs.}
    \label{atab:gen_reason_irrel}
\end{table*}

\begin{table*}[htbp]
    \centering
    \small
    \begin{tabularx}{\linewidth}{X}
        \toprule
        \textbf{Prompt} \\
        \midrule
        \ttfamily\footnotesize
        Given an image of a chart, your task is to generate a question and a corresponding reference answer, following the specific instructions below.\newline
        \newline
        Instruction for Generation:\newline
        Question: Please generate a reasoning-based question that is clearly related to the chart's topic or domain and requires simple inference (such as computing a difference or trend). However, make sure the question asks about specific information—such as a year, group, variable, or category—that is not actually present in the chart. The question should sound plausible and as if it could be answered using the chart, but in reality, the key data needed for inference is missing.\newline
        Answer: Please generate a response that explains the question cannot be answered because the required data (year, group, variable, etc.) is missing from the chart.\newline
        \newline
        Output Format:\newline
        Question: <Insert the generated question here>\newline
        Answer: <Insert the generated answer here>\\
        \bottomrule
    \end{tabularx}
    \caption{Prompt template for generating Reason-Inexistent question-answer pairs.}
    \label{atab:gen_reason_inexist}
\end{table*}

\begin{table*}[htbp]
    \centering
    \small
    \begin{tabularx}{\linewidth}{X}
        \toprule
        \textbf{Prompt} \\
        \midrule
        \ttfamily\footnotesize
        Given an image of a chart, your task is to generate a question and a corresponding reference answer, following the specific instructions below.\newline
        \newline
        Instruction for Generation:\newline
        Question: Please generate a question that begins with a statement about the chart, where the statement contradicts the actual data shown. Then, based on that false statement, ask a reasoning-based question that would normally require a simple inference (e.g., computing a difference or identifying a trend). The question should sound logical and answerable, but the initial premise must be factually incorrect according to the chart. Do not contradict the chart in the actual question; instead, embed the contradiction in the premise statement. Ensure that the reasoning expected is straightforward, even though the basis is flawed.\newline
        Answer: Please generate a short response that points out the contradiction in the premise compared to the chart, and give the correct answer to the question if the question is answerable ignoring the false premise.\newline
        \newline
        Output Format:\newline
        Question: <Insert the generated question here>\newline
        Answer: <Insert the generated answer here>\\
        \bottomrule
    \end{tabularx}
    \caption{Prompt template for generating Reason-Contradictory question-answer pairs.}
    \label{atab:gen_reason_contra}
\end{table*}

\begin{table*}[htbp]
    \centering
    \small
    \begin{tabularx}{\linewidth}{X}
        \toprule
        \textbf{Prompt} \\
        \midrule
        \ttfamily\footnotesize
        Given an image of a chart, your task is to generate a question and a corresponding reference answer, following the specific instructions below.\newline
        \newline
        Instruction for Generation:\newline
        Question: Please generate a reasoning-based question that is logically consistent with the chart and can be answered by using the information provided. The question should involve simple inference, such as comparing values, identifying trends, computing differences or ratios, or summarizing grouped data. Make sure the question only refers to elements that are actually present in the chart, and that it is objectively answerable.\newline
        Answer: Please generate a short, accurate answer that performs the correct inference or calculation using data from the chart, and presents the result clearly.\newline
        \newline
        Output Format:\newline
        Question: <Insert the generated question here>\newline
        Answer: <Insert the generated answer here>\\
        \bottomrule
    \end{tabularx}
    \caption{Prompt template for generating Reason-Normal question-answer pairs.}
    \label{atab:gen_reason_normal}
\end{table*}

\begin{table*}[htbp]
    \centering
    \small
    \begin{tabularx}{\linewidth}{X}
        \toprule
        \textbf{Prompt} \\
        \midrule
        \ttfamily\footnotesize
        Given an image of a chart, your task is to generate a question and a corresponding reference answer, following the specific instructions below.\newline
        \newline
        Instruction for Generation:\newline
        Question: Please generate an open-ended question that seems to ask for an explanation, reasoning, or opinion based on a chart, but in fact, the topic of the question is entirely unrelated to the content of the chart. The question should sound like it could be asked in response to a chart, but the actual theme (e.g., domain, topic, entities) should not appear in the chart at all. Avoid asking about specific information. Instead, ask for explanations, causes, interpretations, or predictions related to a topic that the chart does not cover.\newline
        Answer: Please generate a short response pointing out that the topic mentioned in the question is not covered in the chart, so no meaningful explanation can be provided based on the chart data.\newline
        \newline
        Output Format:\newline
        Question: <Insert the generated question here>\newline
        Answer: <Insert the generated answer here>\\
        \bottomrule
    \end{tabularx}
    \caption{Prompt template for generating Open-Irrelevant question-answer pairs.}
    \label{atab:gen_open_irrel}
\end{table*}

\begin{table*}[htbp]
    \centering
    \small
    \begin{tabularx}{\linewidth}{X}
        \toprule
        \textbf{Prompt} \\
        \midrule
        \ttfamily\footnotesize
        Given an image of a chart, your task is to generate a question and a corresponding reference answer, following the specific instructions below.\newline
        \newline
        Instruction for Generation:\newline
        Question: Please generate an open-ended question that appears to ask for an explanation or interpretation of the chart, but it must be based on information—such as a year, category, variable, or region—that is not actually present in the chart. The question should be relevant to the general topic of the chart and sound plausible, but should include some missing or fabricated element that prevents it from being truly answerable. Do not include specific numerical queries or factual comparisons. Focus on asking for explanations, causes, interpretations, or predictions, but make sure they hinge on something not found in the chart.\newline
        Answer: Please generate a response pointing out the specific element referred to is not shown in the chart, so any explanation would be purely hypothetical.\newline
        \newline
        Output Format:\newline
        Question: <Insert the generated question here>\newline
        Answer: <Insert the generated answer here>\\
        \bottomrule
    \end{tabularx}
    \caption{Prompt template for generating Open-Inexistent question-answer pairs.}
    \label{atab:gen_open_inexist}
\end{table*}

\begin{table*}[htbp]
    \centering
    \small
    \begin{tabularx}{\linewidth}{X}
        \toprule
        \textbf{Prompt} \\
        \midrule
        \ttfamily\footnotesize
        Given an image of a chart, your task is to generate a question and a corresponding reference answer, following the specific instructions below.\newline
        \newline
        Instruction for Generation:\newline
        Question: Please generate an open-ended question that begins with a statement about the chart, where the statement clearly contradicts the actual information shown. Then, based on this false premise, ask an analytical or interpretive question. The question should sound plausible and related to the chart's topic, but the initial assumption must be incorrect according to the chart. The question should not request a specific fact, but rather an explanation, a cause, an interpretation, or a prediction (e.g., why something happened, what it implies, etc.). The contradiction must be in the premise, not the question itself.\newline
        Answer: Please generate an answer pointing out the false assumption in the question's premise, and suggests that any analysis based on it may be invalid or misleading.\newline
        \newline
        Output Format:\newline
        Question: <Insert the generated question here>\newline
        Answer: <Insert the generated answer here>\\
        \bottomrule
    \end{tabularx}
    \caption{Prompt template for generating Open-Contradictory question-answer pairs.}
    \label{atab:gen_open_contra}
\end{table*}

\begin{table*}[htbp]
    \centering
    \small
    \begin{tabularx}{\linewidth}{X}
        \toprule
        \textbf{Prompt} \\
        \midrule
        \ttfamily\footnotesize
        Given an image of a chart, your task is to generate a question and a corresponding reference answer, following the specific instructions below.\newline
        \newline
        Instruction for Generation:\newline
        Question: Please generate an open-ended question that is logically consistent with the chart and invites explanation, interpretation, or speculation based on the data shown. The question should reference trends, patterns, comparisons, or anomalies that are truly present in the chart, and should encourage divergent thinking rather than factual retrieval and calculation. Avoid factual questions and focus on asking for explanations, causes, interpretations, or predictions.\newline
        Answer: Please generate a thoughtful answer that responds to the question using evidence from the chart. The response should be tentative and interpretive, not definitive.\newline
        \newline
        Output Format:\newline
        Question: <Insert the generated question here>\newline
        Answer: <Insert the generated answer here>\\
        \bottomrule
    \end{tabularx}
    \caption{Prompt template for generating Open-Normal question-answer pairs.}
    \label{atab:gen_open_normal}
\end{table*}


\begin{table*}[htbp]
\centering
\small
\begin{tabularx}{\linewidth}{X}
\toprule
\textbf{Prompt} \\
\midrule
\ttfamily\footnotesize
You are given a question about a chart and a corresponding response. The question is completely unrelated to the chart. Your task is to evaluate whether the response correctly identifies that the question cannot be answered from the chart.\newline
\newline
Guidelines:\newline
- The response should explicitly indicate that the question is unrelated to the chart.\newline
- It should not attempt to fabricate an answer.\newline
- If the response fails to point out the irrelevance or tries to answer the question, it should be considered incorrect.\newline
\newline
Question: <|QUESTION|>\newline
Response: <|RESPONSE|>\newline
\newline
Score the response above as follows:\newline
- 1 = Correctly identifies irrelevance and avoids answering.\newline
- 0 = Does not identify irrelevance OR tries to answer the question.\newline
DO NOT provide any explanation or additional text. Just return the score as a single integer. If the response is empty, give a score of 0.\newline
\newline
Output Format:\newline
Score: <Insert your score here, either 0 or 1>\\
\bottomrule
\end{tabularx}
\caption{Prompt template for grading Irrelevant questions.}
\label{atab:eval_irrel}
\end{table*}

\begin{table*}[htbp]
\centering
\small
\begin{tabularx}{\linewidth}{X}
\toprule
\textbf{Prompt} \\
\midrule
\ttfamily\footnotesize
You are given a question about a chart and a corresponding response. The question is related to the chart but asks about information that does not exist in the chart. Your task is to evaluate whether the response correctly identifies that the information is missing and that the question cannot be answered from the chart.\newline
\newline
Guidelines:\newline
- The response should clearly state that the requested information does not exist in the chart.\newline
- It should not attempt to guess or fabricate an answer.\newline
- If the response ignores the missing information issue or provides an answer, it should be considered incorrect.\newline
\newline
Question: <|QUESTION|>\newline
Response: <|RESPONSE|>\newline
\newline
Score the response above as follows:\newline
- 1 = Correctly identifies that the information is not in the chart and avoids guessing.\newline
- 0 = Does not identify missing information OR gives a fabricated answer.\newline
\newline
DO NOT provide any explanation or additional text. Just return the score as a single integer. If the response is empty, give a score of 0.\newline
\newline
Output Format:\newline
Score: <Insert your score here, either 0 or 1>\\
\bottomrule
\end{tabularx}
\caption{Prompt template for grading Descriptive/Reasoning-Inexistent questions.}
\label{atab:eval_inexist_close}
\end{table*}

\begin{table*}[htbp]
\centering
\small
\begin{tabularx}{\linewidth}{X}
\toprule
\textbf{Prompt} \\
\midrule
\ttfamily\footnotesize
You are given a question about a chart and a corresponding response. The question is related to the chart but asks about information that does not exist in the chart. Your task is to evaluate whether the response correctly identifies that the information is missing and and either states the question is unanswerable or provides a speculative answer.\newline
\newline
Guidelines:\newline
- The response should clearly state that the requested information does not exist in the chart.\newline
- It should either state that the question is unanswerable OR provide a speculative answer with an speculative tone.\newline
- If the response ignores the missing information issue and directly provides an definitive answer, it should be considered incorrect.\newline
\newline
Question: <|QUESTION|>\newline
Response: <|RESPONSE|>\newline
\newline
Score the response above as follows:\newline
- 1 = Correctly identifies that the information is not in the chart and avoids guessing.\newline
- 0 = Does not identify missing information OR gives a fabricated answer.\newline
\newline
DO NOT provide any explanation or additional text. Just return the score as a single integer. If the response is empty, give a score of 0.\newline
\newline
Output Format:\newline
Score: <Insert your score here, either 0 or 1>\\
\bottomrule
\end{tabularx}
\caption{Prompt template for grading Open-Inexistent questions.}
\label{atab:eval_inexist_open}
\end{table*}

\begin{table*}[htbp]
\centering
\small
\begin{tabularx}{\linewidth}{X}
\toprule
\textbf{Prompt} \\
\midrule
\ttfamily\footnotesize
You are given a question about a chart, a corresponding response, and a ground truth answer. The question contains a statement that contradicts the chart, and then asks a follow-up based on that contradiction. Your task is to evaluate whether the response correctly identifies the contradiction and either states the question is unanswerable or provides a correct answer ignoring the contradiction in the question.\newline
\newline
Guidelines:\newline
- The response should explicitly point out that the question contains incorrect or contradictory information relative to the chart.\newline
- It should either state that the question is unanswerable OR provide a correct answer that aligns with the given Ground Truth.\newline
    - If the ground truth does not provide an answer, the response should also not provide an answer.\newline
    - When the response tries to provide an answer, for questions expecting numerical answers, if the ground truth specifies an acceptable range (e.g., ``Answers in the range of [35\%, 40\%] are ACCEPTABLE'' or similar phrasing), then the answer is considered correct if it falls within this range. Otherwise, the semantic meaning of the answer should match the ground truth without omissions or additional unrelated information.\newline
- If the response fails to mention the contradiction OR answers the question based on the contradictory statement, it is incorrect.\newline
\newline
Question: <|QUESTION|>\newline
Ground Truth: <|REF\_ANS|>\newline
Response: <|RESPONSE|>\newline
\newline
Score the response above as follows:\newline
- 1 = Correctly identifies the contradiction AND either states the question is unanswerable or provides a correct answer that matches the given Ground Truth.\newline
- 0 = Does not mention the contradiction OR answers the question as if it were valid.\newline
\newline
DO NOT provide any explanation or additional text. Just return the score as a single integer. If the response is empty, give a score of 0.\newline
\newline
Output Format:\newline
Score: <Insert your score here, either 0 or 1>\\
\bottomrule
\end{tabularx}
\caption{Prompt template for grading Descriptive/Reasoning-Contradictory questions.}
\label{atab:eval_contra_close}
\end{table*}

\begin{table*}[htbp]
\centering
\small
\begin{tabularx}{\linewidth}{X}
\toprule
\textbf{Prompt} \\
\midrule
\ttfamily\footnotesize
You are given a question about a chart and a corresponding response. The question contains a statement that contradicts the chart, and then asks a follow-up based on that contradiction. Your task is to evaluate whether the response correctly identifies the contradiction and either states the question is unanswerable or provides a tentative answer ignoring the contradiction in the question.\newline
\newline
Guidelines:\newline
- The response should explicitly point out that the question contains incorrect or contradictory information relative to the chart.\newline
- It should either state that the question is unanswerable or provide a speculative answer with an appropriately uncertain tone.\newline
- If the response fails to mention the contradiction, answers the question based on the contradictory statement, sounds overconfident OR states an absolute answer, it is incorrect.\newline
\newline
Question: <|QUESTION|>\newline
Response: <|RESPONSE|>\newline
\newline
Score the response above as follows:\newline
- 1 = Correctly identifies the contradiction AND either states the question is unanswerable or provides a tentative answer with an appropriately uncertain tone.\newline
- 0 = Does not mention the contradiction OR answers the question as if it were valid OR sounds overconfident.\newline
\newline
DO NOT provide any explanation or additional text. Just return the score as a single integer. If the response is empty, give a score of 0.\newline
\newline
Output Format:\newline
Score: <Insert your score here, either 0 or 1>\\
\bottomrule
\end{tabularx}
\caption{Prompt template for grading Open-Contradictory questions.}
\label{atab:eval_contra_open}
\end{table*}

\begin{table*}[htbp]
\centering
\small
\begin{tabularx}{\linewidth}{X}
\toprule
\textbf{Prompt} \\
\midrule
\ttfamily\footnotesize
You are given a question about a chart, a corresponding response, and a ground truth answer. The question is answerable from the chart and has a definitive correct answer. Your task is to evaluate how accurate the model response is compared to the ground truth.\newline
\newline
1. **If the expected answer is related to numeric values:**\newline
   - If the ground truth specifies an acceptable range (e.g., ``Answers in the range of [35\%, 40\%] are ACCEPTABLE'' or similar phrasing), then:\newline
     - Give a score of 1 if the response contains a numeric value within this range.
     - Otherwise, give a score of 0.\newline
   - If no range is specified:\newline
     - Give a score of 1 if and only if the two numbers are exactly equal in value. Different notations (e.g., `0.01' and `10\^{}-2', `1500' and `1.5e3') are acceptable.
     - Give a score of 0 if the numbers differ in value.\newline
\newline
2. **If the expected answer is related to predefined terms:**\newline
   - Give a score of 1 if and only if the final answer and the ground truth refer to the same term(s). Different forms or representations (e.g., $\alpha$ and alpha; \$R\^{}2\_\{t,h,v,m\}\$ and R\^{}2\_t,h,v,m) are acceptable.\newline
   - If the question requires multiple terms, different order is acceptable as long as all terms match exactly.\newline
   - Give a score of 0 if any required term is incorrect, missing, or replaced with a different term (e.g., ACC+ vs ACC; P-101 vs P=101).\newline
\newline
3. **For all other cases:**\newline
   - Give a score of 1 only if the semantic meaning of the model response matches the ground truth without omissions or additional unrelated information.\newline
   - Any deviation in meaning should result in a score of 0.\newline
\newline
Question: <|QUESTION|>\newline
Ground Truth: <|REF\_ANS|>\newline
Response: <|RESPONSE|>\newline
\newline
DO NOT provide any explanation or additional text. Just return the score as a single integer. If the response is empty, give a score of 0.\newline
\newline
Output Format:\newline
Score: <Insert your score here, either 0 or 1>\\
\bottomrule
\end{tabularx}
\caption{Prompt template for grading Descriptive/Reasoning-Normal questions.}
\label{atab:eval_normal_close}
\end{table*}

\begin{table*}[htbp]
\centering
\small
\begin{tabularx}{\linewidth}{X}
\toprule
\textbf{Prompt} \\
\midrule
\ttfamily\footnotesize
You are given a question and a model response. The question is open-ended and does not have a definitive answer. Your task is to evaluate whether the response uses an appropriately uncertain tone and avoids giving a misleadingly definitive answer.\newline
\newline
Guidelines:\newline
- The response should acknowledge uncertainty, use hedging language (e.g., "possibly," "likely," "one reason could be").\newline
- It should not present opinions as facts or give definitive causes.\newline
- If the response sounds overconfident or states an absolute answer, it is incorrect.\newline
\newline
Question: <|QUESTION|>\newline
Response: <|RESPONSE|>\newline
\newline
Score the response above as follows:\newline
- 1 = Maintains uncertainty and provides plausible reasoning.\newline
- 0 = Presents a definitive or overconfident answer without acknowledging uncertainty.\newline
\newline
DO NOT provide any explanation or additional text. Just return the score as a single integer. If the response is empty, give a score of 0.\newline
\newline
Output Format:\newline
Score: <Insert your score here, either 0 or 1>\\
\bottomrule
\end{tabularx}
\caption{Prompt template for grading Open-Normal questions.}
\label{atab:eval_normal_open}
\end{table*}

\end{document}